\documentclass{article}

\usepackage[preprint]{neurips_2019}

\usepackage[utf8]{inputenc} 
\usepackage[T1]{fontenc}    
\usepackage{hyperref}       
\usepackage{url}            
\usepackage{booktabs}       
\usepackage{amsfonts}       
\usepackage{nicefrac}       
\usepackage{microtype}      

\usepackage{graphicx}
\usepackage{wrapfig}
\usepackage{subcaption}
\usepackage{booktabs} 
\usepackage{times} 
\usepackage{amsmath} 
\usepackage{amssymb}  
\usepackage{bbm}
\usepackage{mathtools}
\DeclareMathOperator*{\argmax}{\arg\!\max}

\usepackage{textcomp}
\usepackage{color}
\usepackage{soul}
\usepackage{nth}
\usepackage{multirow}
\usepackage{siunitx}
\usepackage{paralist}

\newcommand{\norm}[1]{\left\lVert#1\right\rVert}

\title{Scheduled Intrinsic Drive: A Hierarchical Take on Intrinsically Motivated Exploration}

\author{
	Jingwei Zhang\thanks{Authors contributed equally.} ,
	Niklas Wetzel$^{*}$,
	Nicolai Dorka$^{*}$,
	Joschka Boedecker,
	Wolfram Burgard\\
	\texttt{\{zhang,wetzel,dorka,jboedeck,burgard\}@cs.uni-freiburg.de}\\
	University of Freiburg, Germany\\
}

\begin{document}

\maketitle

\begin{abstract}
Exploration in sparse reward reinforcement learning remains an open challenge.
Many state-of-the-art methods use intrinsic motivation to complement the sparse extrinsic reward signal,
giving the agent more opportunities to receive feedback during exploration.
Commonly these signals are added as bonus rewards,
which results in a mixture policy that neither conducts exploration nor task fulfillment resolutely.
In this paper,
we instead learn separate intrinsic and extrinsic task policies and schedule between these different drives
to accelerate exploration and stabilize learning.
Moreover,
we introduce a new type of intrinsic reward denoted as \textit{successor feature control} (\textbf{SFC}),
which is general and not task-specific.
It takes into account statistics over complete trajectories and thus differs from previous methods that only use local information to evaluate intrinsic motivation.
We evaluate our proposed \textit{scheduled intrinsic drive} (\textbf{SID}) agent using three different environments with pure visual inputs:
VizDoom,
DeepMind Lab
and DeepMind Control Suite.
The results show a substantially improved exploration efficiency with SFC and the hierarchical usage of the intrinsic drives.
A video of our experimental results can be found at \url{https://youtu.be/b0MbY3lUlEI}.

\end{abstract}

%

\section{Introduction}
\label{sec:introduction}
Reinforcement learning (RL) agents learn on evaluative feedback (reward signals) instead of instructive feedback (ground truth labels),
which takes the process of automating the development of intelligent problem-solving agents one step further \citep{sutton2018reinforcement}.
With deep networks as powerful function approximators bringing traditional RL into high-dimensional domains,
deep reinforcement learning (DRL) has shown great potential \citep{mnih2015human,mnih2016asynchronous,schulman2017proximal,horgan2018distributed}.
However,
the success of DRL often relies on carefully shaped dense extrinsic reward signals.
Although shaping extrinsic rewards can greatly support the agent in finding solutions and shortening the interaction time,
designing such dense extrinsic signals often requires substantial domain knowledge,
and calculating them typically requires ground truth state information,
both of which is hard to obtain in the context of robots acting in the real world.
When not carefully designed,
the reward shape could sometimes serve as bias
or even distractions and could potentially hinder the discovery of optimal solutions.
More importantly,
learning on dense extrinsic rewards goes backwards on the progress of reducing supervision
and could prevent the agent from taking full advantage of the RL framework.

In this paper,
we consider terminal reward RL settings,
where a signal is only given when the final goal is achieved.
When learning with only an extrinsic terminal reward indicating the task at hand,
intelligent agents are given the opportunity to potentially discover optimal solutions even out of the scope of the well established domain knowledge.

However,
in many real-world problems defining a task only by a terminal reward means that the learning signal can be extremely sparse.
The RL agent would have no clue about what task to accomplish until it receives the terminal reward for the first time by chance.
Therefore in those scenarios guided and structured exploration is crucial,
which is where intrinsically-motivated exploration \citep{oudeyer2008can,schmidhuber2010formal} has recently gained great success \citep{pathakICMl17curiosity,burda2018exploration}.
  Most commonly in current state-of-the-art approaches, an intrinsic reward is added as a reward bonus to the extrinsic reward.
  Maximizing this combined reward signal, however, results in a mixture policy that neither acts greedily with regard to extrinsic reward maximization nor to exploration.
  Furthermore, the non-stationary nature of the intrinsic signals could potentially lead to unstable learning on the combined reward.
  In addition, current state-of-the-art methods have been mostly looking at local information calculated out of 1-step lookahead for the estimation of the intrinsic rewards, e.g.
  one step prediction error \citep{pathakICMl17curiosity}, or
  network distillation error of the next state \citep{burda2018exploration}.
  Although those intrinsic signals can be propagated back to earlier states with temporal difference (TD) learning,
  it is not clear that this results in optimal long-term exploration. 
We seek to address the aforementioned issues as follows:
\begin{enumerate}
  \item
  We propose a hierarchical agent \textit{scheduled intrinsic drive} (\textbf{SID}) that focuses on one motivation at a time:
  It learns two separate policies which maximize the extrinsic and intrinsic rewards respectively.
  A high-level scheduler periodically selects to follow either
  the extrinsic or the intrinsic
  policy to gather experiences. 
  Disentangling the two policies allows the agent to faithfully conduct either pure exploration or pure extrinsic task fulfillment.
  Moreover,
  scheduling (even within an episode) inexplicitely increases the behavior policy space exponentially,
  which drastically differs from previous methods where the behavior policy could only change slowly due to the incremental nature of TD learning.
  \item
  We introduce \textit{successor feature control} (\textbf{SFC}), a novel intrinsic reward that is based on the concept of successor features. This feature representation characterizes states through the features of all its successor states
  instead of looking at local information only.
  This implicitly makes our method
  temporarily extended,
  which enables more structured and far-sighted exploration that is crucial in exploration-challenging environments.
\end{enumerate}

We note that both the proposed intrinsic reward SFC and the hierarchical exploration framework SID are without any task-specific components,
and can be incorporated into existing DRL methods with minimal computation overhead.
We present experimental results in three sets of environments,
evaluating our proposed agent in the domains of visual navigation and control from pixels,
as well as its capabilities of finding optimal solutions under distraction.

\section{Related Work}
\label{sec:relatedWorks}


\paragraph{Intrinsic Motivation and Auxiliary Tasks}

Intrinsic motivation can be defined as agents conducting actions purely out of the satisfaction of its internal rewarding system rather than the extrinsic rewards \citep{oudeyer2008can,schmidhuber2010formal}.
There exist various forms of intrinsic motivation 
and they have achieved substaintial improvement in guiding exploration
for DRL,
in tasks where extrinsic signals are sparse or missing altogether.

\citep{pathakICMl17curiosity} proposed to evaluate curiosity,
one of the most widely used kinds of intrinsic motivation,
with the $1$-step prediction error of the features of the next state made by a forward dynamics model.
Their ICM module has been shown to work well in visual domains including first-person view navigation.
Since ICM is potentially susceptible to stochastic transitions \citep{burda2018large},
\citet{burda2018exploration} propose as a reward bonus the error of predicting the features of the current state output by a randomly initialized fixed embedding network.
Another form of curiosity,
learning progress or the change in the prediction error,
has been connected to count-based exploration via a pseudo-count \citep{bellemare2016unifying,ostrovski2017count} and has also been used as a reward bonus.
\citet{savinov2018episodic} propose to train a reachability network,
which gives out a reward based on whether the current state is reachable within a certain amount of steps from any state in the current episode.
Similar to our proposed SFC,
their intrinsic motivation is related to choosing states that could lead to novel trajectories.
However we note that the reachability reward bonus captures the novelty of states with regard to the current episode,
while our proposed SFC reward implicitly captures statistics over the full distribution of policies that have been followed,
since the successor features are learned using states sampled from all past experiences.


Auxiliary tasks have been proposed for learning more representative and distinguishable features.
\citet{mirowski2016learning} add depth prediction and loop closure prediction as auxiliary tasks for learning the features.
\citet{jaderberg2016reinforcement} learn separate policies for maximizing pixel changes (pixel control) and activating units of a specific hidden layer (feature control).
However,
their proposed UNREAL agent never follows those auxiliary policies as they are only used to learn more suitable features for the main extrinsic task.



\paragraph{Hierarchical RL}

Various HRL approaches have been proposed
\citep{kulkarni2016hierarchical,bacon2017option,vezhnevets2017feudal,
krishnan2017ddco}.
In the context of intrinsic motivation,
feature control \citep{jaderberg2016reinforcement} has been adopted into a hierarchical setting \citep{dilokthanakul2017feature},
in which options are constructed for altering given features.
However,
they report that a flat policy trained on the intrinsic bonus achieves similar performance to the hierarchical agent.

Our hierarchical design is perhaps inspired mostly by the work of \citet{riedmiller2018learning}.
Unlike other HRL approaches that try to learn a set of options \citep{sutton1999between} to construct the optimal policy,
their proposed SAC agent aims to learn one flat policy that maximizes the extrinsic reward.
While SAC schedules between following the extrinsic task and a set of pre-defined auxiliary tasks such as
maximizing touch sensor readings or translation velocity,
in this paper
we investigate scheduling between the extrinsic task and intrinsic motivation that is general and not task-specific.

\paragraph{Successor Representation}

The successor representation (SR) was first introduced to improve generalization in TD learning \citep{dayan1993improving}.
While previous works extended SR to the deep setting for better generalized navigation and control algorithms across similar environments and changing goals \citep{kulkarni2016deep,barreto2017successor,zhang2017deep},
we focus on its temporarily extended property to accelerate exploration.
%

SR has also been investigated under the options framework.
\citet{machado2017eigenoption,tomar*2019successor} evaluate successor features with random policies to discover bottlenecks or landmarks based on the clustering of such features.
Options are then learned to navigate to those sub-goals.
However,
it remained unclear if the options framework would help in sparse exploration setups.



When using SR to measure the intrinsic motivation,
the most relevant work to ours is that of \citet{machado2018count}.
They also design a task-independent intrinsic reward based on SR,
however they rely on the concept of count-based exploration and propose a reward bonus,
that vastly differs from ours.
We will present our proposed method in the next section.

\section{Methods}
\label{sec:methods}

We use the RL
framework for learning and decision-making under uncertainty.
It is formalized by Markov decision processes (MDPs) defined by the tuple $\langle\mathcal{S},\mathcal{A},p,r,\gamma \rangle$.
At time step $t$ the agent samples an action $a\in\mathcal{A}$ according to policy $\pi(\cdot|s)$,
which depends on its current state $s\in\mathcal{S}$.
The agent receives a scalar reward $r\in\mathbb{R}$ and transits to the next state $s'\in\mathcal{S}$.
The distribution of the corresponding state, action and reward process $(S_t,A_t,R_{t+1})$ is determined by the distribution of the initial state $S_0$, the transition operator $p$ and the policy $\pi$.
The goal of the agent is to find a policy
that maximizes the expectation of the sum of discounted rewards
$\sum_{k=0}^T \gamma^k R_{t+k+1}$.
We seek to speed up learning in sparse reward RL,
where the reward signal is uninformative for almost all transitions. We set the focus on terminal reward scenarios, where the agent only receives a single reward of $+1$ for successfully accomplishing the task and $0$ otherwise.

We will first introduce our proposed intrinsic reward \textit{successor feature control} (\textbf{SFC})
(\ref{sec:sfc2},\ref{sec:sfc3}),
then present our proposed hierachical framework for accelerating intrinsically motivated exploration,
which we denote as \textit{scheduled intrinsic drive} (\textbf{SID}) (Sec.\ref{sec:sid1},\ref{sec:sid2}).

\subsection{Successor Distance Metric}
\label{sec:sfc2}

In order to encode long-term statistics into the design of intrinsic rewards for far-sighted exploration,
we build on the formulation of \textit{successor represention} (SR),
which introduces a temporarily extended view of the states.
\citet{dayan1993improving} introduced the idea of representing a state $s$ by the occupancies of all other states from a process starting in $s$ following a fixed policy $\pi$,
where the occupancies denote the average number of time steps the state process stays in each state per episode.
\textit{Successor features} (SF) extend the concept to an arbitrary feature embedding
$\phi:\mathcal{S}\rightarrow \mathbb{R}^m$.
For a fixed policy $\pi$ and embedding $\phi$ the SF is defined by the $|m|$-dimensional vector
\begin{align}
    \psi_{\pi,\phi}(s)
&:=
    \mathbb{E}_{\pi} \left[ \sum_{t=0}^\infty \gamma^{t} \phi(S_{t}) \Big| S_0 = s \right]\label{eq:SF}.
\end{align}
Analogously,
the SF represent the average discounted feature activations, when starting in $s$ and following $\pi$.
They can be learned by \emph{temporal difference} (TD) updates
\begin{equation}
\psi_{\pi,\phi}(S_t)
\leftarrow
\psi_{\pi,\phi}(S_t)+
\alpha\Big[
{\phi}({S_{t+1}}) +
\gamma \psi_{\pi}(S_{t+1}) -
\psi_{\pi}(S_{t}) \Big].
\label{eq:learn-sf}
\end{equation}



SF have several interesting properties which make them appealing as a basis for an intrinsic reward signal:
\begin{inparaenum}[1)]
\item
They can be learned
even in the absence of extrinsic rewards and without learning a transition model and therefore combine advantages of model-based and model-free RL
\citep{stachenfeld2014design}.
\item They can be learned via computationally efficient TD.
\item They capture the expected feature activations for complete episodes.
Therefore they contain information even of
spatially and temporarily
distant states
which might help for effective far-sighted exploration.
\end{inparaenum}
Given the discussion,
we introduce
the successor distance (SD) metric that measures the distance between states by the similarity of their SF
\begin{align}
    d_{\pi,\phi}(s,s'):=||\psi_{\pi,\phi}(s)- \psi_{\pi,\phi}(s')||_2.
    \label{eq:dist}
\end{align}
This definition bases on a well know approach in distance metric learning that defines distances by $d_W(x_1,x_2)^2=(x_1-x_2)^TW(x_1-x_2)$.
This can be seen by
identifying the feature embedding with the $m\times|\mathcal{S}|$ dimensional matrix $\Phi(i,j):=\phi(s_j)_i$ and SR with the $|\mathcal{S}|\times |\mathcal{S}|$ matrix $\Psi_\pi(i,j):=\psi_\pi(s_i,s_j)$.
Then for $W=\Psi_\pi^T\Phi^T\Phi\Psi_\pi$ the distance measures $d_\pi$ and $d_W$ are equal.
$W$ is symmetric and positive semi-definite (Eq.\ref{eq:dist}) thus $d_W$ defines a pseudometric.


\begin{wrapfigure}{r}{0.24\textwidth}
    \vspace{-15pt}
    \begin{center}
        \includegraphics[width=0.23\textwidth]{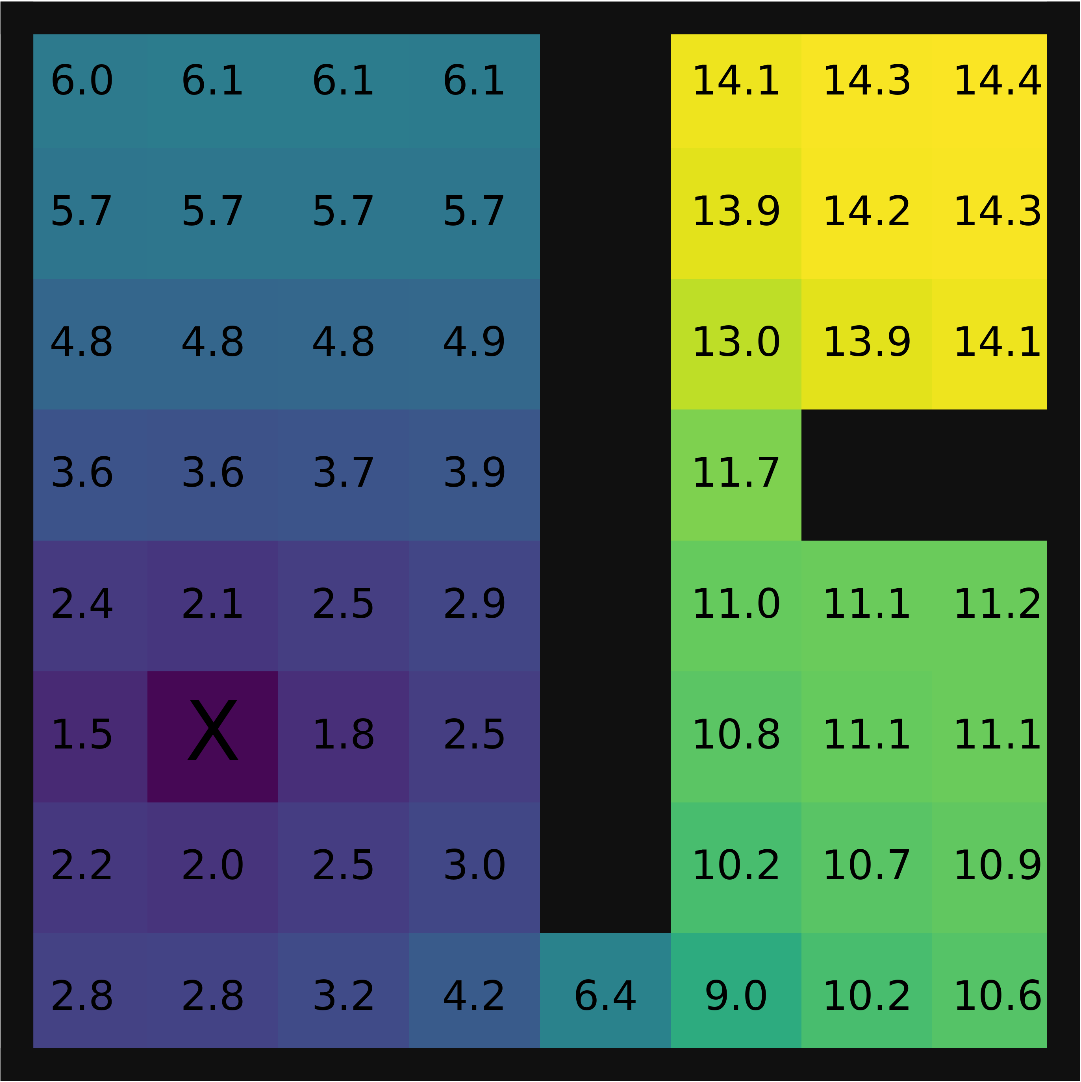}
    \end{center}
    \vspace{-10pt}
    \caption{SD}
    \label{fig:anchor_dist}
    \vspace{-15pt}
\end{wrapfigure}
In Fig.\ref{fig:anchor_dist},
we illustrate the SD
in a grid world with three rooms.
Each value indicates the SD from the corresponding state to a fixed anchor state marked by $\times$,
with the
SF learned using a random walk ($\gamma=0.98$, $\phi$ one-hot encoding).
In this case,
the SD correlates roughly to the length of the shortest path from each state to the anchor.
Notably is that the SD increases substantially when crossing rooms.
When starting from the anchor state with a random policy,
it is relatively unlikely for the agent to enter the other two rooms;
thus for a pair of states with a fixed spatial distance,
their SD is higher when they locate in different rooms than in the same room.
So the SD also captures the connectivity in the state space.

\subsection{Successor Feature Control}
\label{sec:sfc3}
\begin{wrapfigure}{r}{0.24\textwidth}
    \vspace{-15pt}
    \begin{center}
        \includegraphics[width=0.23\textwidth]{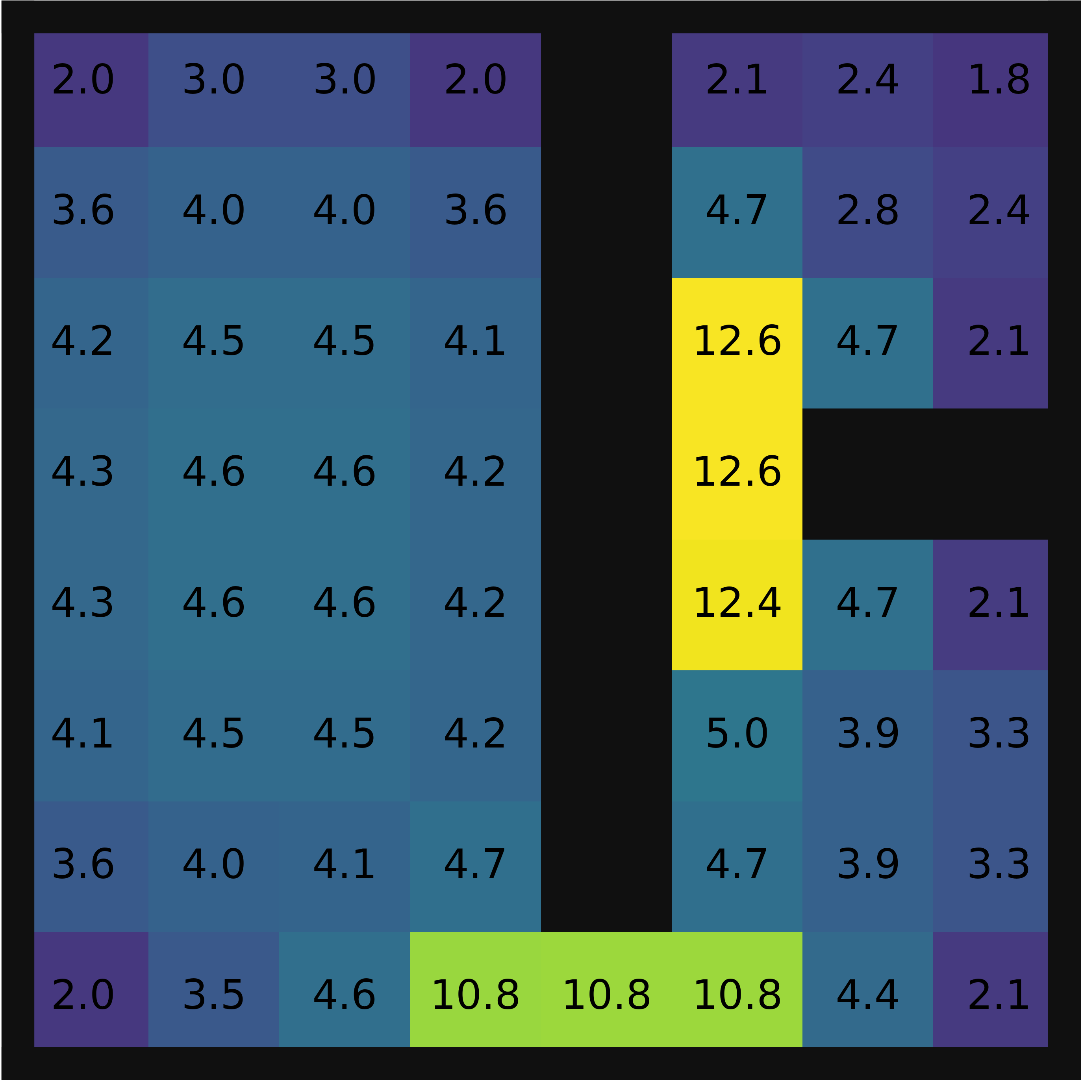}
    \end{center}
    \vspace{-10pt}
    \caption{SFC}
    \vspace{-15pt}
    \label{fig:sfc}
\end{wrapfigure}
Using this metric to evaluate the intrinsic motivation,
one choice would be to use the SD to a fixed anchor state as the intrinsic reward,
which depends heavily on the anchor position.
Even when a sensible choice for the anchor can be found, e.g. the initial state of an episode,
the SDs of distant states from the anchor assimilate.
To circumvent this,
we define the intrinsic reward \textit{successor feature control} (SFC) as the squared SD of
a pair of consecutive states
in Eq.\ref{eq:sfc-1}.
A high SFC reward indicates a big change in the future feature activations when $\pi$ is followed.
We argue this big change
is a strong indicator of bottleneck states,
since in bottlenecks a minor change in the action selection can lead to a vastly different trajectory being taken.
This is especially true for highly stochastic policies.
Fig.\ref{fig:sfc} shows that those highly rewarding states under SFC and the true bottlenecks agree,
which can be very valuable for exploration
\citep{lehnert2018value}.
\begin{align}
    R^{\text{sfc}}_{t+1}
:=
    \norm{\psi_{\pi,\phi}(S_{t+1})-\psi_{\pi,\phi}(S_{t})}^2_2.
    \label{eq:sfc-1}
\end{align}
Another valuable property of SFC is that
it adapts in very meaningful ways that lead to efficient non-stationary exploration policies,
when the transitions gathered by a policy maximizing the SFC reward is used to update the SF itself.
Intuitively the SFC reward and the SD update pull in opposite directions.
This can be seen by looking at the SD before and after updating the SF
with a transition from $s$ to $s'$.
Taking this transition effectively reduced the SD between $s$ and $s'$,
because the SF of $s$ are pushed to the direction of the SF of $s'$
(the successors of $s'$ are the successors of $s$ as well).
Therefore the SFC of a transition would be reduced after this transition is taken,
discouraging the agent
to take the same transition again.
Thus SFC has similarities with count-based exploration bonuses, but has a straight forward extension to deep learning.

\subsection{Scheduled Intrinsic Drive}
\label{sec:sid1}
When learning optimal value functions or optimal policies via TD or policy gradient with deep function approximators,
optimizing with algorithms such as gradient descent means that the policy would only evolve incrementally:
It is necessary that the TD-target values do not change drastically over a short period of time in order for the gradient updates to be meaningful.
The common practice of utilizing a target network in off-policy DRL \citep{mnih2015human} stabilizes the update but in the meanwhile making the policy adapt even more incrementally over each step.

But intrinsically motivated exploration,
or exploration in general,
might benefit from an opposite treatment of the policy update.
This is because the intrinsic reward is non-stationary by nature,
as well as the fact that the exploration policy should reflect the optimal strategy corresponding to the current stage of learning,
and thus is also non-stationary.

With the commonly adopted way of using intrinsic reward as a bonus to the extrinsic reward and train a mixture policy on top,
exploration would be a balancing act between the incrementally updated target values for stable learning and the dynamically adapted intrinsic signals for efficient exploration.
Moreover,
neither the extrinsic nor the intrinsic signal is followed for an extended amout of time.

Therefore,
we propose to address this issue with a hierarchical approach that by design has slowly changing target values while still allowing drastic behavior changes.
The idea is to learn not a single,
but multiple policies,
with each one optimizing on a different reward function.
To be more specific, 
we assume to have $N$ tasks $\mathbb{T}\in\mathcal{T}$
(e.g. $N=2$ and $\mathcal{T}=\{\mathbb{T}_{\text{E}},\mathbb{T}_{\text{I}}\}$ where $\mathbb{T}_{\text{E}}$ denotes the extrinsic task and $\mathbb{T}_{\text{I}}$ the intrinsic task)
defined by $N$ reward functions (e.g. $R_{\text{E}}$ and $R_{\text{I}}$)
that share the state and action space.
The optimal policy for each of these $N$ different MDPs can be learned with arbitrary off-policy DRL algorithms.
During each episode,
a high-level scheduler periodically selects a policy for the agent to follow to gather experiences,
and each policy is trained with all experiences collected following those $N$ different policies.
The overall learning objective is to maximize the extrinsic reward
$
    \mathbb{E}_{\omega(\mathbb{T}|S_t)}
    \mathbb{E}_{\pi_{\mathbb{T}}(A_t|S_t)}
    \left[
        q_{\mathbb{T}_E}(S_t, A_t | A_t\sim\pi_{\mathbb{T}}(\cdot|S_t))
    \right]
$
($\omega$: the macro-policy of the scheduler).

By allowing the agent to follow one motivation at a time,
it is possible to have a pool of $N$ different behavior policies without creating unstable targets for off-policy learning.
By scheduling $M$ times even during an episode,
we inexplicitely increase the behavior policy space by exponential to $N^M$ for a single episode.
We investigated several types of high-level schedulers,
however,
none of them consistently outperforms a random one.
We suspect the reason why a random scheduler already performs very well under the SID framework,
is that a highly stochastic schedule can be beneficial to make full use of the big behavior policy space.
We present the different scheduler choices we tested in Appendix \ref{app:scheduler},
and leave more sophisticated scheduler design to future work.
Moreover,
disentangling the extrinsic and intrinsic policy strictly separates stationary and non-stationary behaviors,
and the different sub-objectives would each be allocated with its own interaction time,
such that extrinsic reward maximizaton and exploration do not distract each other.

\subsection{Algorithm Implementation}
\label{sec:sid2}
Our proposed method can be combined with an any approach that allows off-policy learning.
This section describes an instantiation of the SID framework when using Ape-X DQN as a basic off-policy DRL algorithm \cite{horgan2018distributed}
with SFC as the intrinsic reward, which we used for all experiments. For details see Appendix \ref{app:implementation}.
The algorithm is composed of:
\begin{itemize}
	\item A Q-Net $\{ \theta_{\varphi}, \theta_{\text{E}}, \theta_{\text{I}} \}$:
	Contains a shared embedding $\theta_{\varphi}$ and two Q-value output heads $\theta_{\text{E}}$ (extrinsic) and $\theta_{\text{I}}$ (intrinsic).
	\item A SF-Net $\{ \theta_{\phi}, \theta_{\psi} \}$:
	Contains an embedding $\theta_{\phi}$ and a successor feature head $\theta_{\psi}$. $\theta_{\phi}$ is initialized randomly and kept fixed during training.
	The output of SF-Net is used to calculate the SFC intrinsic reward (Eq.\ref{eq:sfc-1}).
	\item A high-level scheduler:
	Instantiated in each actor,
	selects which policy to follow (extrinsic or intrinsic) after a fixed number of environment steps (max episode length$/M$). The sheduler randomly picks one of the tasks with equal probability.
	\item $N$ parallel actors ($N=8$):
	Each actor instantiates its own copy of the environment,
	periodically copies the latest model from the learner.
	We learn from $K$-step targets ($K=5$),
	so each actor at each environment step stores $(s_{t-K}, a_{t-K}, \sum_{k=1}^K\gamma^{k-1}r_{t-K+k}, s_{t})$ into a shared replay buffer. 
	Each actor will act according to either the extrinsic or the intrinsic policy based on the current task selected by its scheduler.
	\item A learner:
	Learns the Q-Net ($\theta_{\text{E}}$ and $\theta_{\text{I}}$ are learned with the extrinsic and intrinsic reward respectively) and the SF-Net from samples (Eq.\ref{eq:learn-sf}) from the same shared replay buffer,
	which contains all experiences collected from following different policies.
\end{itemize}
We depict this algorithm instance in Appendix Fig.\ref{fig:alg}.

\section{Experiments}
\label{sec:experiments}

\begin{figure}[b]
    \begin{subfigure}{0.2\columnwidth}
        \centering
        \vspace{10pt}
        \includegraphics[width=\columnwidth]{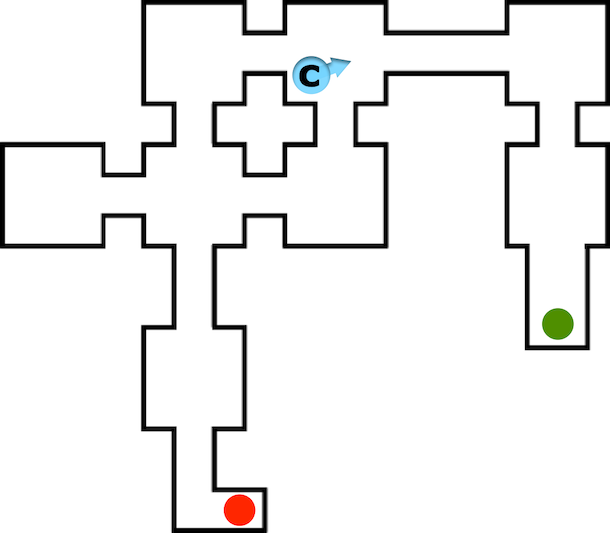}
        \caption{MyWayHome}
        \label{fig:mwh_layout}
    \end{subfigure}
    \begin{subfigure}{0.7\columnwidth}
        \centering
        \includegraphics[width=\columnwidth]{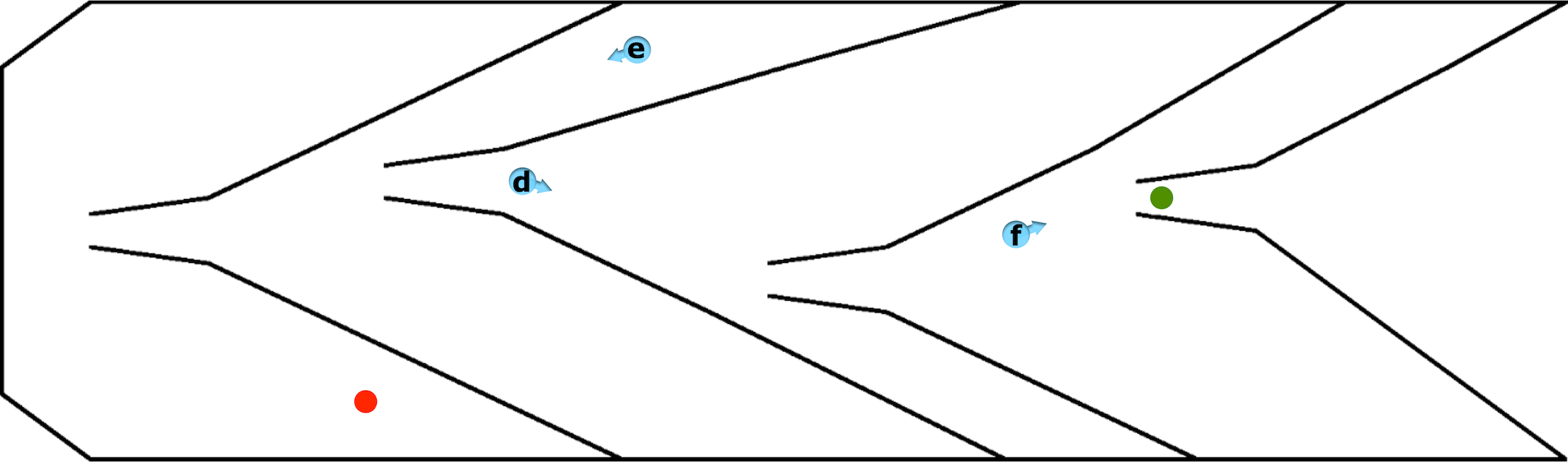}
        \caption{FlytrapEscape}
        \label{fig:fly_layout}
    \end{subfigure}
    \begin{subfigure}{0.23\columnwidth}
        \centering
        \includegraphics[width=\columnwidth]{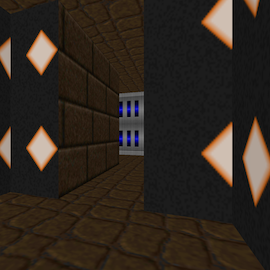}
        \caption{Corridor.}
        \label{fig:mwh_screen1}
    \end{subfigure}
    \hfill
    \begin{subfigure}{0.23\columnwidth}
        \centering
        \includegraphics[width=\columnwidth]{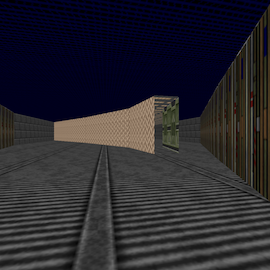}
        \caption{Exit.}
        \label{fig:fly_screen1}
    \end{subfigure}
    \begin{subfigure}{0.23\columnwidth}
        \centering
        \includegraphics[width=\columnwidth]{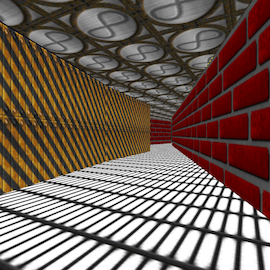}
        \caption{Wing.}
        \label{fig:fly_screen2}
    \end{subfigure}
    \begin{subfigure}{0.23\columnwidth}
        \centering
        \includegraphics[width=\columnwidth]{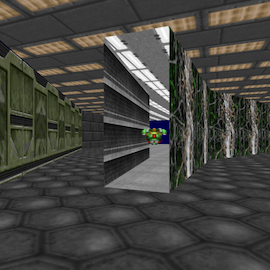}
        \caption{Goal.}
        \label{fig:fly_screen3}
    \end{subfigure}

    \caption{
        VizDoom environments we evaluated on.
        \ref{fig:mwh_layout} and \ref{fig:fly_layout} show the top-down views of MyWayHome and FlytrapEscape with the same downscaling ratio,
        with red dots marking the starting locations,
        green dots indicating the goal locations;
        \ref{fig:mwh_screen1} and \ref{fig:fly_screen1} to \ref{fig:fly_screen3} show exemplary first-person views captured from the marked poses (blue dots with arrows) from those two maps respectively.}
     \label{fig:fly_env}
\end{figure}

%

We evaluate our proposed intrinsic reward SFC and the hierarchical framework of intrinsic motivation SID in three sets of simulated environments:
VizDoom \citep{kempka2016vizdoom},
DeepMind Lab \citep{beattie2016deepmind} and
DeepMind Control Suite \citep{tassa2018deepmind}.
Throughout all experiments,
agents receive as input only raw pixels with no additional domain knowledge or task specific information.
We mainly compare the following agent configurations:
\textbf{M}: Ape-X DQN with 8 actors, train with only the extrinsic main task reward;
\textbf{ICM}: train a single policy with the ICM reward bonus \citep{pathakICMl17curiosity};
\textbf{RND}: train a single policy with the RND reward bonus \citep{burda2018exploration};
\textbf{Ours}: with our proposed SID framework, schedule between following the extrinsic main task policy and the intrinsic policy trained with our proposed SFC reward.

We carried out an ablation study, where we compare the performance of an agent with intrinsic and extrinsic reward summed up, to the corresponding SID agent for each intrinsic reward type (ICM, RND, SFC). We present the plots and discussions in Appendix \ref{app:ablation}.

For the intrinsic reward normalization and the scaling for the extrinsic and intrinsic rewards we do a parameter sweep
for each environment (Appendix \ref{app:rewardnorm})
and choose the best setting for each agent.
We notice that our scheduling agent is much less sensitive to different scalings than agents with added reward bonus.
Since our proposed SID setup requires an off-policy algorithm to learn from experiences generated by following different policies,
we implement all the agents under the Ape-X DQN framework \cite{horgan2018distributed}.
After a parameter sweep we set the number of scheduled tasks per episode to $M=8$ for our agent in all experiments,
meaning each episode is divided into up to $8$ sub-episodes,
and for each of which either the extrinsic or the intrinsic policy is sampled as the behavior policy.
Appendix \ref{app:implementation} and \ref{app:training}
contain additional information about experimental setups and model training details.





\subsection{VizDoom: Sparse Navigation}

\begin{figure}[t]
    \centering
    \begin{subfigure}[t]{0.49\textwidth}
        \includegraphics[width=\columnwidth]{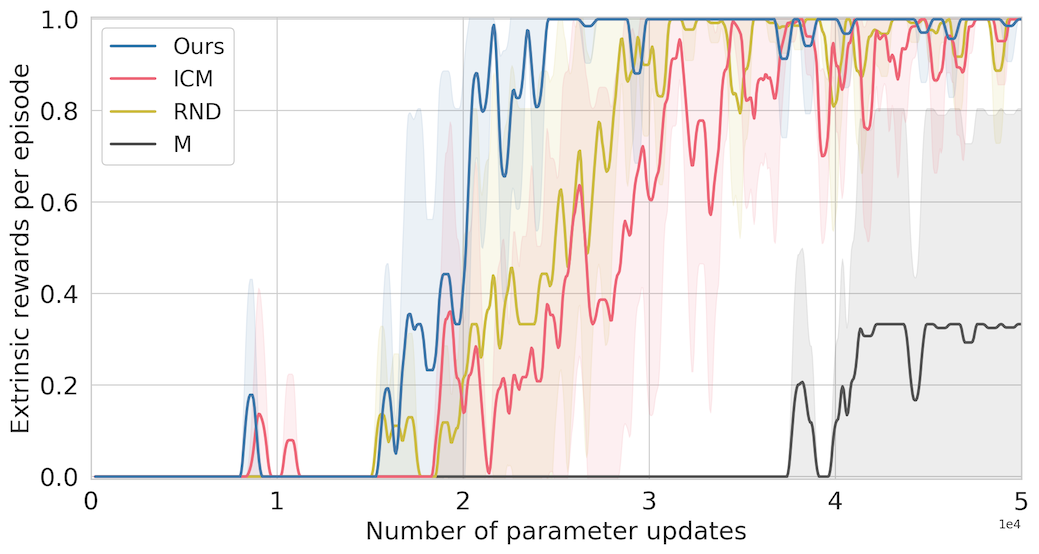}
        \label{fig:mwh_plot}
    \end{subfigure}
    \hfill
    \begin{subfigure}[t]{0.49\textwidth}
        \includegraphics[width=\columnwidth]{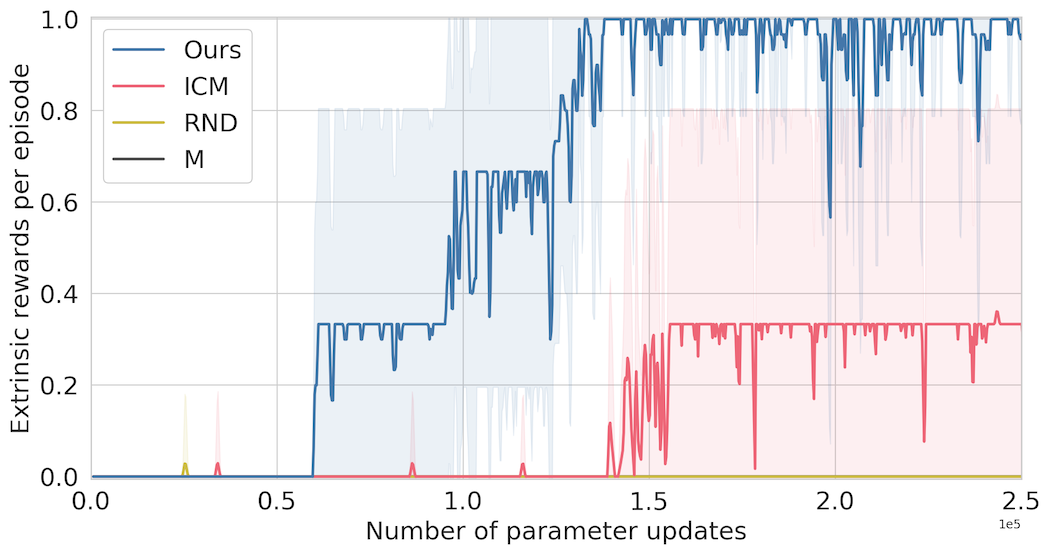}
        \label{fig:fly_plot}
    \end{subfigure}
    \vspace{-10pt}
        \caption{
        Extrinsic rewards per episode obtained in MyWayHome (left) and FlytrapEscape (right).
        Each plot shows the mean with $\pm1$ standard deviation over 3 non-tuned random seeds.
        }
    \label{fig:doom_plots}
\end{figure}

We start by verifying our implementation of the baseline algorithms in "DoomMyWayHome" which was previously used in several state-of-the-art intrinsic motivation papers \citep{pathakICMl17curiosity,savinov2018episodic}.
The agent needs to navigate based only on first-person view visual inputs through 8 rooms connected by corridors (Fig.\ref{fig:mwh_layout}),
each with a distinct texture (Fig.\ref{fig:mwh_screen1}).
The experimental results are shown in Fig.\ref{fig:doom_plots} (left).
Since our basic RL algorithm is doing off-policy learning, it has relatively decent random exploration capabilities.
We see that the M agent is able to solve the task sometimes
without any intrinsically generated motivations, but that all intrinsic motivation types help to solve the task more reliably and speed up the learning. Our method solve the task the fastest, but also ICM and RND learn to reach the goal reliably and efficiently.

We wanted to test the agents on a more difficult VizDoom map
where structured exploration would be of vital importance.
We thus designed a new map which scales up the navigation task of MyWayHome.
Inspired by how flytraps catch insects,
we design the layout of the rooms in a geometrically challenging way that escaping from one room to the next with random actions is extremely unlikely.
We show the layout of MyWayHome (Fig.\ref{fig:mwh_layout}) and FlytrapEscape (Fig.\ref{fig:fly_layout}) with the same downscaling ratio.
The maze consists of 4 rooms separated by V-shaped walls pointing inwards the rooms.
The small exists of each room is located at the junction of the V-shape,
which is extremely difficult to maneuver into without a sequence of precise movements.
As in the MyWayHome task,
in each episode,
the agent starts from the red dot shown in Fig.\ref{fig:fly_layout} with a random orientation.
An episode terminates if the final goal is reached and the agent will receive a reward of $+1$,
or if a maximum episode steps of \num[group-separator={,}]{10000}
(2100 for MyWayHome)
is reached.
The task is to escape the fourth room.

The experimental results on FlytrapEscape are shown in Fig.\ref{fig:doom_plots} (right).
Neither M nor RND manages to learn any useful policies.
ICM solves the task in sometimes,
while we can clearly observe that our method efficiently explores the map and reliably learns how to navigate to the goal.

\begin{wrapfigure}{r}{0.5\textwidth}
    \vspace{-15pt}
    \begin{center}
        \includegraphics[width=0.49\textwidth]{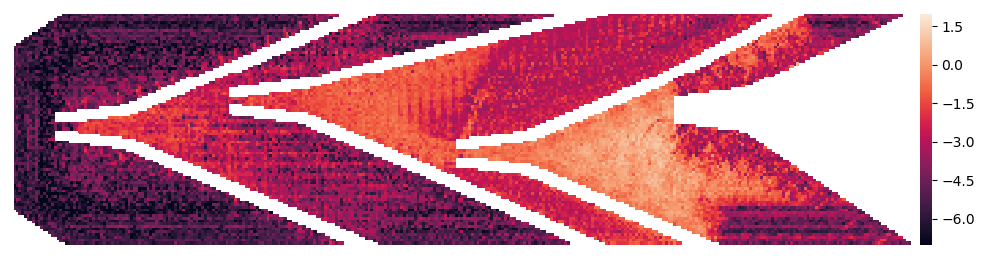}
    \end{center}
    \vspace{-10pt}
    \caption{
    Projection of the SFs (Appendix \ref{app:fly_sf}).
    }
    \label{fig:fly_sf}
    \vspace{-10pt}
\end{wrapfigure}
As an additional evaluation,
we visualize the SF of an Ours agent which successfully learned to navigate to the goal (Fig.\ref{fig:fly_sf}).
We can see that the SD from each coordinate to the starting position tends to grow as the geometric distance increases,
especially for those that locate on the
pathways leading to later rooms.
This shows that the learned SD and the geometric distance are in good agreement and that the SF are learned as expected.
Furthermore,
we observe big intensity changes around the bottlenecks (the room entries) in the heatmap,
which also supports the hypothesis that SFC leads the agent to bottleneck states.
We believe this is the first time that SF are shown to behave in a first-person view environment as one would expect from its definition.
The evolution of the SF over time is shown in the attached video.

\subsection{DeepMind Lab: Exploration under Distraction}
\begin{wrapfigure}{r}{0.29\textwidth}
    \vspace{-15pt}
    \begin{center}
        \includegraphics[width=0.28\textwidth]{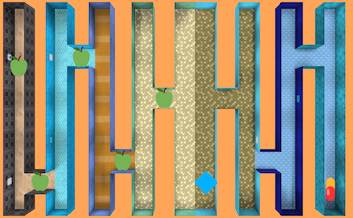}
    \end{center}
    \vspace{-10pt}
    \caption{AppleDistractions.}
    \label{fig:apple_layout}
    \vspace{-10pt}
\end{wrapfigure}
In the second experiment,
we set out to evaluate if the agents would be able to reliably collect the faraway big reward in the presence of small nearby distractive rewards.
For this experiment we use the 3D visual navigation simulator of DeepMind Lab   \citep{beattie2016deepmind}.
We constructed a challenging level "AppleDistractions"
(Fig.\ref{fig:apple_layout}) with a maximum episode length of $1350$.
In this level,
the agent starts in the middle of the map (blue square) and can follow either of the two corridors.
Each corridor has multiple sections and each section consists of two dead-ends and an entry to next section.
Each section has different randomly generated floor and wall textures.
One of the corridors (left) gives a small reward of $0.05$ for
each apple collected,
while the other one (right) contains a single big reward of $1$ at the end of its last section.
The optimal policy would be to go for the single faraway big reward.
But since the small apple rewards are much closer to the spawning location of the agent,
the challenge here is to still explore other areas sufficiently often so that the optimal solution could be recovered.

\begin{figure}[t]
    \begin{subfigure}[t]{0.49\textwidth}
        \includegraphics[width=\columnwidth]{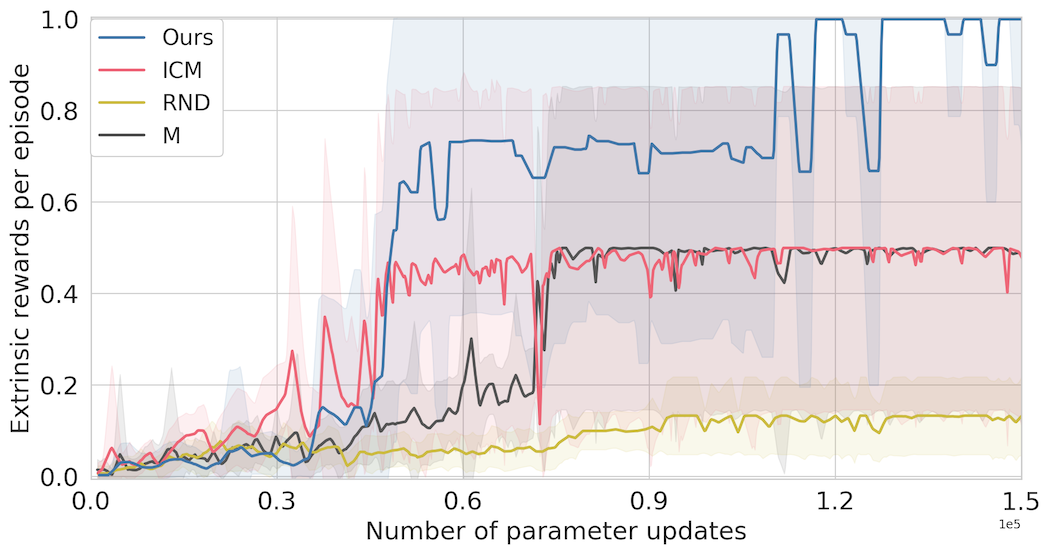}
        \label{fig:lab_plot}
    \end{subfigure}
    \hfill
    \begin{subfigure}[t]{0.49\textwidth}
        \includegraphics[width=\columnwidth]{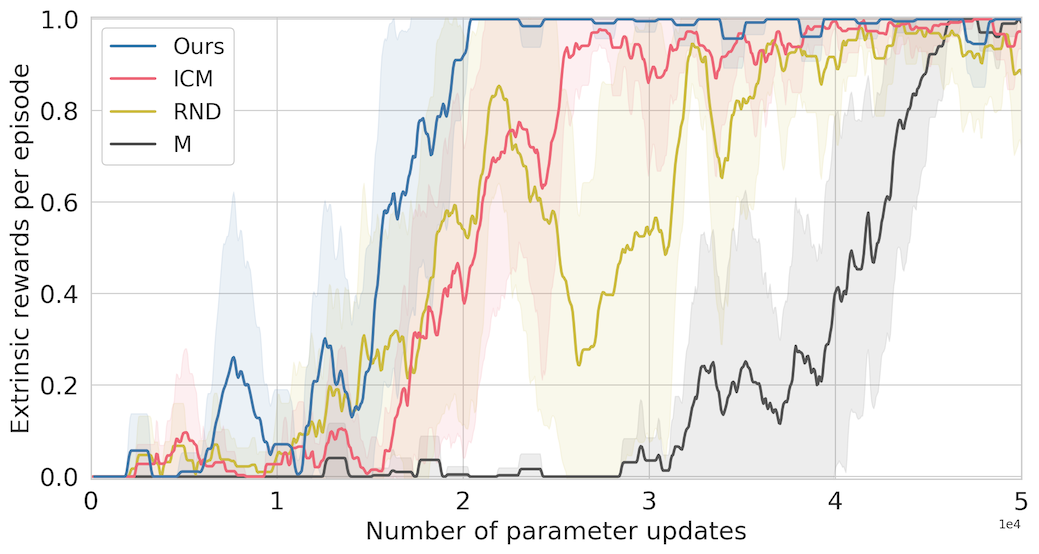}
        \label{fig:cart_plot}
    \end{subfigure}
    \vspace{-10pt}
        \caption{
        Extrinsic rewards per episode obtained in AppleDistractions (left) and Cartpole (right).
        Each plot shows the mean with $\pm1$ standard derivation over 3 non-tuned random seeds.
        Left:
        Each agent is evaluated on the same $3$ sets of random floor and wall textures,
        with $3$ non-tuned environment seeds.
        In the ablation study (Appendix \ref{app:ablation}) the SID variant ourperms the reward bonus variant of each of the $3$ types of intrinsic rewards.
        Right:
        Ours also outforms all baseline agents in the very different domain of classic control from pixels,
        which shows the general applicability of our proposed agent.
        }
        \label{fig:lab_cartpole_plots}
\end{figure}

The results are presented in Fig.\ref{fig:lab_cartpole_plots} (left).
Ours received on average the highest rewards and is the only method that learns to navigate to the large reward in every run.
The baseline methods get easily distracted by the small short-term rewards and do not reliably learn to navigate away from the distractions.
With a separate policy for intrinsic motivation the agent can for some time interval completely "forget" about the extrinsic reward and purely explore,
since it does not get distracted by the easily reachable apple rewards and can efficiently learn to explore the whole map.
In the meanwhile the extrinsic policy can simultaneously learn from the new experiences and might learn about the final goal discovered by the exploration policy.
This highlights a big advantage of scheduling over bonus rewards,
that it reduces the probability of converging to bad local optimums.
In Appendix \ref{app:ablation} we further showed that SID is generally applicable and also helps ICM and RND in this task.

\subsection{DeepMind Control Suite: Classic Control from Pixels}
\begin{wrapfigure}{r}{0.2\textwidth}
    \vspace{-15pt}
    \begin{center}
        \includegraphics[width=0.2\textwidth]{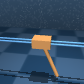}
    \end{center}
    \vspace{-10pt}
    \caption{Cartpole.}
    \label{fig:cartpole_screen}
    \vspace{-30pt}
\end{wrapfigure}
To show that our methods can be used in domains other than first-person visual navigation,
we evaluate on the classic control task "carpole: swingup\_sparse" (DeepMind Control Suite \cite{tassa2018deepmind}),
using third-person view images as inputs (Fig.\ref{fig:cartpole_screen}).
The pole starts pointing down and the agent receives a single terminal reward of $+1$ for swinging up the unactuated pole using only horizontal forces on the cart.
Additional details are presented in Appendix \ref{app:dmcontrol}.
The results are shown in Fig.\ref{fig:lab_cartpole_plots} (right).
Compared to the previous tasks,
this task is easy enough to be solved without intrinsic motivation,
but we can see also that all intrinsic motivation methods significantly reduce the interaction time.
Ours still outperforms other agents even in the absence of clear bottlenecks which shows its general applicability,
but since the task is relatively less challenging for exploration,
the performance gain is not as substantial as the previous experiments.

%

\section{Conclusion}
\label{sec:conclusion}

In this paper,
we investigate an alternative way of utilizing intrinsic motivation for exploration in DRL.
We propose a hierarchical agent SID that schedules between following extrinsic and intrinsic drives.
Moreover,
we propose a new type of intrinsic reward SFC that is general and evaluates the intrinsic motivation based on longer time horizons.
We conduct experiments in three sets of environments
and show that both our contributions
SID and SFC
help greatly in improving exploration efficiency.

We consider many possible research directions that could stem from this work,
including designing more efficient scheduling strategies,
incorporating several intrinsic drives (that are possibly orthogonal and complementary) instead of only one into SID,
testing our framework in other control domains such as manipulation,
and extending our evaluation onto real robotics systems.


\subsubsection*{Acknowledgments}

The authors would like to thank Martin Riedmiller, Daniel Büscher, Gabriel Kalweit, Artemij Amiranashvili and Max J. Argus for frutiful discussions and their valuable feedbacks.

\clearpage
\bibliography{zhang19neurips}
\bibliographystyle{unsrtnat}


\clearpage
\appendix

\section{Appendix: Ablation Study}
\label{app:ablation}

\begin{figure}[t]
	\centering
		\includegraphics[width=0.7\columnwidth]{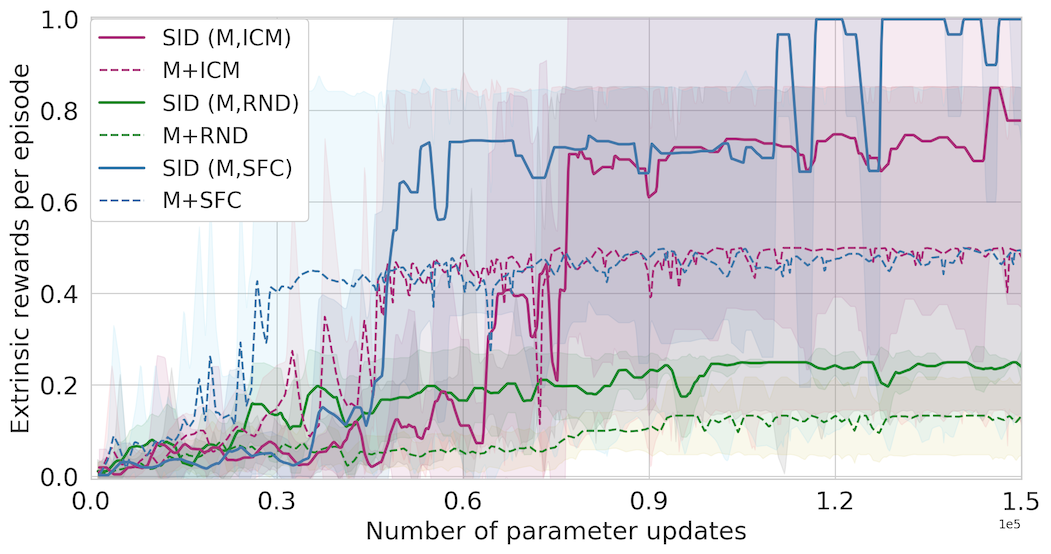}
		\label{fig:lab_ablation}
	\caption{Ablation study results for AppleDistractions.
	\label{fig:apple_ablation}
	}
\end{figure}
We have conducted ablation studies for all the three sets of environments to investigate
the influence of scheduling on our proposed method,
whether other reward types can benefit from scheduling too,
and whether environment specific differences exist.

We compare the performance of the following agent configurations:
\begin{itemize}
	\item Three reward bonus agents
	\textbf{M+ICM},
	\textbf{M+RND},
	\textbf{M+SFC}:
	\\The agent receives the intrinsic reward of
	ICM \citep{pathakICMl17curiosity},
	RND \citep{burda2018exploration} or
	our proposed SFC
	respectively as added bonus to the extrinsic main task reward and trains a mixture policy on this combined reward signal.
	We note that the \textbf{M+ICM} and \textbf{M+RND} agent in this section corresponds to the \textbf{ICM} and \textbf{RND} agent in all other sections respectively.
	\item Three SID agents
	\textbf{SID (M, ICM)},
	\textbf{SID (M, RND)},
	\textbf{SID (M, SFC)}:
	\\The agent schedules between following the extrinsic main task policy and the intrinsic policy trained with the ICM, RND or our proposed SFC reward respectively.
\end{itemize}
	We note that the \textbf{SID (M, SFC)} agent in this section corresponds to the \textbf{Ours} agent in all other sections.

In Fig.\ref{fig:apple_ablation},
we present the ablation study results for AppleDistractions.
Our SID(M, SFC) agent received on average the highest rewards.
Furthermore,
we see that scheduling helped both ICM and SFC to find the goal and not settle for the small rewards,
and SID also helps improve the performance of RND.
The respective reward bonus counterparts of the three SID agents were more be attracted to the small nearby rewards.
This behavior is expected:
By scheduling,
the intrinsic policy of the SID agent is assigned with its own interaction time with the environment,
during which it could completely "forget" about the extrinsic rewards.
The agent then has a much higher probability of discovering the faraway big reward, thus escaping the distractions of the nearby small rewards.
Once the intrinsic policy collects these experiences of the big reward,
the extrinsic policy can immediately learn from those since both policies share the same replay buffer.

\begin{figure}[t]
	\centering
	\begin{subfigure}[t]{0.49\textwidth}
		\includegraphics[width=\columnwidth]{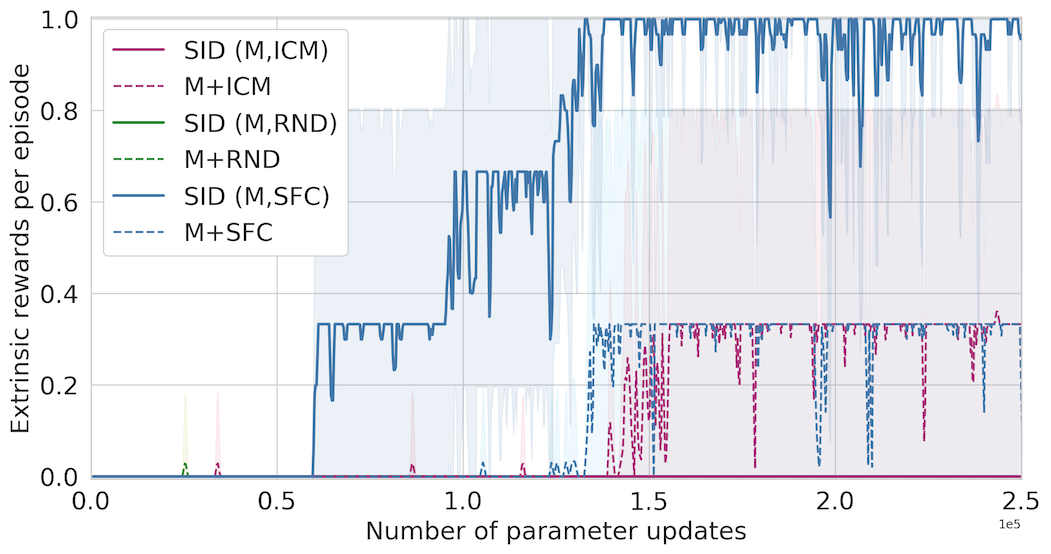}
	\end{subfigure}
	\hfill
	\begin{subfigure}[t]{0.49\textwidth}
		\includegraphics[width=\columnwidth]{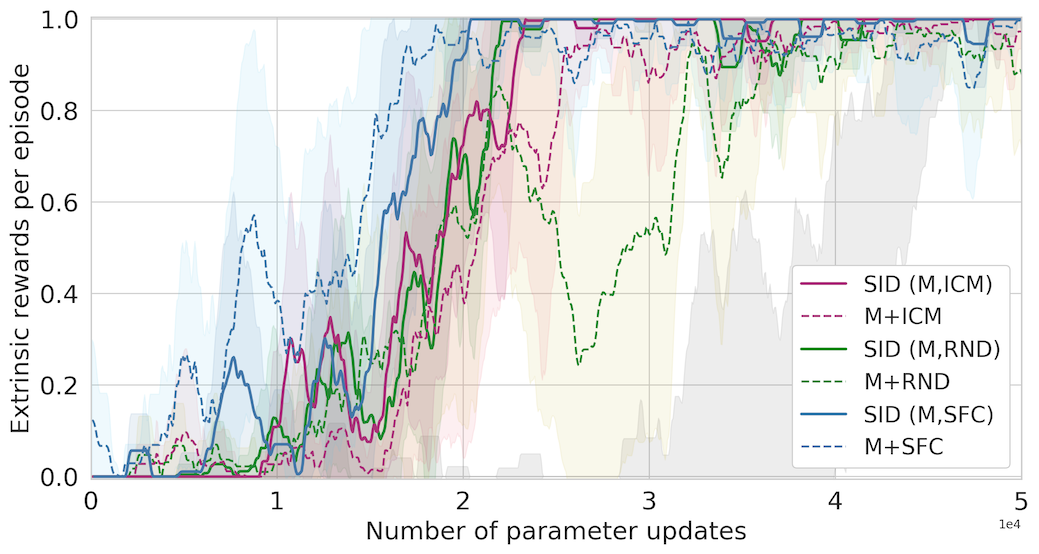}

	\end{subfigure}
\caption{Ablation study results for FlytrapEscape (left) and Cartpole (right).
}
			\label{fig:cartpole_ablation}

\end{figure}

In Fig.\ref{fig:cartpole_ablation} (left),
we present the ablation study results for FlytrapEscape.
The agents with the ICM component perform poorly.
Only $1$ run of M+ICM learned to navigate to the goal,
while the scheduling agent SID(M,ICM) did not solve the task even once.
But for the two SFC agents,
the scheduling greatly improves the performance.
Although the reward bonus agent M+SFC was not successful in every run,
the SID(M,SFC) agent solved the FlytrapEscape in $3$ out of $3$ runs.
We hypothesize the reason for the superior performance of SID(M,SFC) compared to M+SFC could be the following:
Before seeing the final goal for the first time,
the M+SFC agent is essentially learning purely on the SFC reward,
which is equivalent to the intrinsic policy of the scheduling SID(M,SFC) agent.
Since SFC might preferably go to bottleneck states as the difference between the SF of the two neighboring states are expected to be relatively larger for those states .
Since the extrinsic policy is doing random exploration before receiving any reward signal,
it could be a good candidate to explore the next new room from the current bottleneck state onwards.
Then the SFs of the new room will be learned when it is being explored,
which would then guide the agent to the next bottleneck regions.
Thus the SID(M,SFC) agent could efficiently explore from bottleneck to bottleneck,
while the M+SFC agent could not be able to benefit from the two different behaviors under the extrinsic and intrinsic rewards and could oscillate around bottleneck states.
On the other hand,
sheduling did not help ICM or RND.
A reason could be
that ICM or RND is not especially attracted by bottleneck states so it does not help exploration if the agent spends half of the time acting randomly as the extrinsic policy had no reward yet to learn from.
Also since the FlytrapEscape environment is extremely exploration-challenging,
the temporally extended view of our proposed SFC might of vital importance to guide efficient exploration.

In Fig.\ref{fig:cartpole_ablation} (right),
we present the ablation study results for Cartpole.
We can observe that SID helps to improve the performance of both ICM and RND.
As for SFC,
although the reward bonus agent learns a bit faster than the SID agent,
we note that actually all the three SID agent converge to more stable policies,
while the reward bonus agents tend to oscillate around the optimal return.

\section{Appendix: Implementation Details}
\label{app:implementation}

\begin{figure}[t]
	\center
	\includegraphics[width=0.8\columnwidth]{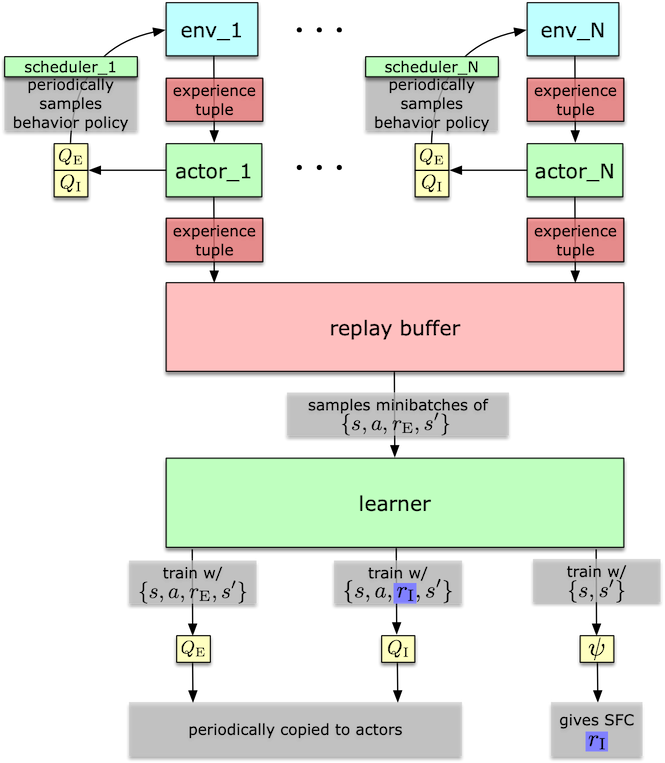}
	\caption{
		Flow diagram of the algorithm implementation (Sec.\ref{sec:sid2}).
	}
	\label{fig:alg}
\end{figure}

This section describes implementation details and design choices. The backbone of our algorithm implementation is presented in Section \ref{sec:sid2} and visualized in Fig. \ref{fig:alg}.

\subsection{Ape-X DQN}
Since our algorithm requires an off-policy learning strategy,
and in consideration for faster learning and less computation overhead,
we use the state-of-the-art off-policy algorithm Ape-X DQN \cite{horgan2018distributed} with the  $K$-step target ($K=5$) for bootstraping without off-policy correction
\begin{align*}
y_t
=
\sum_{k=1}^{k=K}\gamma^{k-1}R_{t+k} +
\gamma^{K}\max q(s_{t+K},\argmax_{a'}q(s_{t+K},a';\theta^{-});\theta),\label{eq:targets}
\end{align*}
where $\theta^{-}$ denotes the target network parameters.

We chose the number of actors the be the highest the hardware supported, which was 8. To adapt the $\epsilon$ settings from the $360$ actors in the Ape-X DQN to our setting of $N=8$ actors,
we set a fixed $\epsilon_i$ for each actor $i\in \{1,\dots,8\}$ as
\begin{align}
\epsilon_i
&=
\epsilon^{1 + \frac{(i-1)\frac{360}{N}}{360-1} \alpha},
\end{align}
where $\alpha=7$ and $\epsilon=0.4$ are set as in the original work.


\subsection{Prioritized Experience Replay}
For computational efficiency,
we implement our own version of the prioritized experience replay.
We split the replay buffer into two, with size of $40,000$ and $10,000$.
Every transition is pushed to the first one, while in the second one only transitions are pushed on which a very large TD-error is computed. We store a running estimate of the mean and the standard deviation of the TD-errors and if for a transition the error is larger than the mean plus two times the standard deviation, the transition is pushed. In the learner a batch of size $128$ consists of $96$ transitions drawn from the normal replay buffer and $32$ are drawn from the one that stores transition with high TD-error, which as a result have relatively seen a higher chance of being picked.

\subsection{Successor Feature Learning}
We note that previous works for learning the deep SF have included an auxiliary task of reconstruction on the features $\phi$ \cite{kulkarni2016hierarchical,zhang2017deep}, 
while in this work we investigate learning $\psi$ without this extra reconstruction stream.
Instead of adapting the features $\phi$ while learning the successor features $\psi$,
we fix the randomly initialized $\phi$. 
This design follows the intuition that since SF ($\psi$) estimates the expectation of features ($\phi$) under the transition dynamics and the policy being followed,
more stable learning of the SF could be achieved if the features are kept fixed.

The SF are learned from the same replay buffer as for training the Q-Net.
Since our base algorithm is $K$-step Ape-X,
and we follow the memory efficient implementation of the replay buffer as suggested in Ape-X,
we only have access to $K$-step experience tuples ($K=5$) for learning the SF. Therefore we calculate the intrinsic reward by applying the canonical extension of the SFC reward formulation (Eq.\ref{eq:sfc-1}) to $K$-step transitions
\begin{align}
R^{\text{sfc}}_{t+K}
&=
\norm{\psi_{\pi,\phi}(S_{t+K})-\psi_{\pi,\phi}(S_{t})}^2_2.
\label{eq:sfc-K}
\end{align}
The behaviour policy $\pi$ associated with the SF is not given explicitly, but since the SF are learned from the replay buffer via TD learning, it is a mixture of current and past behaviour policies from all actors.

\subsection{Reward Normalization}
\label{app:rewardnorm}
Most network parameters are shared for estimating the expected  discounted return of the intrinsic and extrinsic rewards. The scale of the rewards has a big influence on the scale of the gradients for the network parameters. Hence, it is important that the rewards are roughly on the same scale, otherwise effectively different learning rates are applied.
The loss of the network comes from the regression on the Q-values, which approximate the expected return. So our normalization method aims to bring the discounted return of both tasks into the same range.
To do so we first normalize the intrinsic rewards by dividing them by a running estimate of their standard deviation.
We also keep a running estimate of the mean of this normalized reward and denote it $r_I'$.  Since every time step an intrinsic reward is received we estimate the discounted return via the geometric series.
We scale the extrinsic task reward that is always in $\{0,1\}$ with
$\eta \frac{r_I'}{1-\gamma_I}$, where $\gamma_I$ is the discount rate for the intrinsic reward. Furthermore, $\eta$ is a hyperparameter which takes into account that for Q-values from states more distant to the goal the reward is discounted with the discount rate for the extrinsic reward depending on how far away that state is. In our experiments we set $\eta = 3$.

We did the same search for hyperparameters and normalization technique for all algorithms that include an intrinsic reward and found out that the procedure above works best for all of them.
The algorithms were evaluated on the FlytrapEscape.
For $\eta$ we tried the values in $\{0.3,1,3,10\}$.
We also tried to not normalize the rewards and just scale the intrinsic reward. To scale the intrinsic reward we tried the values $\{0.001,0.01,0.1,1\}$. However, we found that as the scale of the intrinsic rewards is not the same over the whole training process this approach does not work well. We also tried to normalize the intrinsic rewards by dividing it by a running estimate of its standard deviation and then scale this quantity with a value in $\{0.01,0.1,1\}$.

\subsection{Model Architecture}
\begin{figure}[t]
	\center
	\includegraphics[width=0.6\columnwidth]{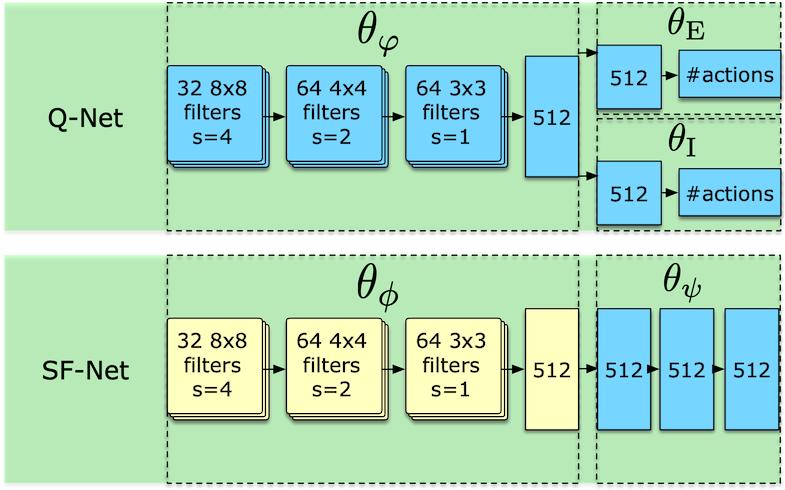}
	\caption{
		Model architecture for the \textbf{SID (M, SFC)} agent.
		Components with color yellow are randomly intialized and not trained during learning.
	}
	\label{fig:model}
	\vspace{-14pt}
\end{figure}

We use the same model architecture as depicted in Fig. \ref{fig:model} across all $3$ sets of experiments.

ReLU activations are added after every layer except for the last layers of each dashed blocks in the above figure.
For the experiments with the ICM \citep{pathakICMl17curiosity},
we added BatchNorm \citep{ioffe2015batch} before activation for the embedding of the ICM module following the original code released by the authors.
Code is implemented in pytorch \citep{paszke2017automatic}.


\section{Appendix: Training Details}
\label{app:training}
We use a batch size of $128$ for all experiments the  Adam optimizer \citep{kingma2014adam} with a learning rate of $1e-4$.

For all experiments we used a stack of 4 consecutive, preprocessed observations as states.

For the first-person view experiments in VizDoom and DeepMind Lab,
we use an action repetition of $4$,
while for the classic control experiment we did not apply action repetition.
In the text,
we only refer to the actual environment steps
(e.g. before divided by 4).

\subsection{Environment Settings: VizDoom}
\label{app:doom}

The VizDoom environment produces $320\times240$ RGB images as observations.
In a preprocessing step, we downscaled the images to $84\times84$ pixels and converted them to grayscale.

For FlytrapEscape, we adopted the action space settings from the MyWayHome task.
The action space was given by the following 5 actions:
\textit{TURN\_LEFT, TURN\_RIGHT, MOVE\_FORWARD, MOVE\_LEFT,	MOVE\_RIGHT}

\subsection{Environment Settings: DeepMind Lab}
\label{app:lab}

\begin{figure}[t]
	\centering
	\begin{subfigure}{0.3\columnwidth}
		\includegraphics[width=\columnwidth]{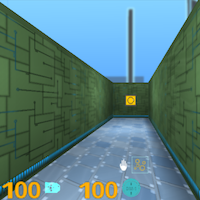}
		\caption{Dead end.}
		\label{fig:apple_screen1}
	\end{subfigure}
	\begin{subfigure}{0.3\columnwidth}
		\includegraphics[width=\columnwidth]{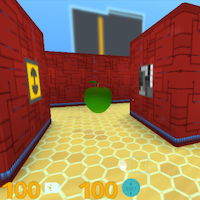}
		\caption{Entry.}
		\label{fig:apple_screen2}
	\end{subfigure}
	\begin{subfigure}{0.3\columnwidth}
		\includegraphics[width=\columnwidth]{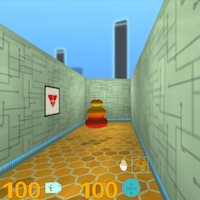}
		\caption{Goal.}
		\label{fig:apple_screen3}
	\end{subfigure}
	\caption{
		Exemplary first-person view observations captured in the AppleDistractions environment.
	}
	\label{fig:apple_env}
\end{figure}

We setup the DmLab environment to produce $84\times84$ RGB images as observations. In Fig.\ref{fig:apple_env} we show examplary observations of AppleDistractions. We preprocessed the images by converting the observations to grayscale.

For a given enviroment seed,
textures for each segment of the maze are generated at random.

We used the predefined DmLab actions from \cite{espeholt2018impala}. The action space was given by the following 8 actions (no shooting setting): \textit{Forward, Backward, Strafe Left, Strafe Right, Look Left, Look Right, Forward+Look Left, Forward+Look Right}.


\begin{figure}[b]
	\centering
	\begin{subfigure}{0.3\columnwidth}
		\includegraphics[width=\columnwidth]{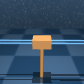}
		\caption{Start}
	\end{subfigure}
	\begin{subfigure}{0.3\columnwidth}
		\includegraphics[width=\columnwidth]{imgs/cartpole/screen}
		\caption{Swingup}
	\end{subfigure}
	\begin{subfigure}{0.3\columnwidth}
		\includegraphics[width=\columnwidth]{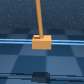}
		\caption{Goal}
	\end{subfigure}
	\caption{
		Exemplary observations captured in the Cartpole environment.
	}
	\label{fig:cartpole_observations}
\end{figure}

\subsection{Environment Settings: DeepMind Control Suite}
\label{app:dmcontrol}
We conducted the experiments for the classic control task on the 'Cart-pole' domain with the 'swingup\_sparse' task provided by the DeepMind Control Suite.
Since our agents needs a discrete action space and the control suite only provides continuous action spaces, we discretized the single action dimension. The set of actions was
\{-0.5, -0.25, 0, 0.25, 0.5\}. We configured the environment to produce 84$\times$84 RGB pixel-only observations from the 1st camera, which is the only predefined camera that shows the full cart and pole at all times. We further convert the images to grey-scale and stack four consecutive frames as input to our network. The episode length was 200 environment steps.


\subsection{Infrastructure}
To generate our results we used two machines that run Ubuntu 16.04.
Each machine has 4 GeForce Titan X (Pascal) GPUs. On one machine we run 4 experiments in parallel, each experiment on a separate GPU.

\section{Appendix: Successor Distance Visualization}
\label{app:fly_sf}


Fig.\ref{fig:fly_sf} visualizes the top-down projection of the SFs of a SID(M,SFC) agent after it learned how to navigate to the goal on FlytrapEscape (Fig.\ref{fig:fly_layout}).
For the purpose of visualization we discretized the map into $85\times330$ grids and position the trained agent SID(M,SFC) at each grid,
then computed the successor features $\psi$ for that location for each of the $4$ orientations $(\ang{0},\ang{90},\ang{180},\ang{270})$,
which resulted in a $4\times512$ matrix.
We then calculated the $l2$-difference of this matrix with a $4\times512$ vector containing the successor features of the starting position with the $4$ different orientations.
Shown in $\log$-scale.

As an additional evaluation,
we visualize the SF of our agent that successfully learned to navigate to the goal (Fig.\ref{fig:fly_sf}).
We can see that the SD from each coordinate to the starting position tends to grow as the geometric distance increases,
especially for those that locate on the
pathways leading to later rooms.
This shows that the learned SD and the geometric distance are in good agreement and that the SF are learned as expected.
Furthermore,
we see big intensity changes around the bottlenecks (the room entries) in the heatmap,
which also supports the hypothesis that SFC leads the agent to bottleneck states.
We believe this is the first time that SF are shown to behave in a first-person view environment as one would expect from its definition.

\section{Appendix: Scheduler Designs}
\label{app:scheduler}

We investigated three types of high-level schedulers:
\begin{itemize}
	\item Random scheduler:
	Sample a task from uniform distribution every task steps.
	\item Switching scheduler:
	Sequentially switches between extrinsic and intrinsic task.
	\item Macro-Q Scheduler:
	Learn a scheduler that learns with macro actions and from sub-sampled experience tuples.
	In each actor,
	we keep an additional local buffer that stores $N+1$ subsampled experiences:
	$ \{ s_{t-Nm}, \ldots, s_{t-2m}, s_{t-m}, s_t \} $.
	Then at each environment step,
	Besides the $K$-step experience tuple mentioned above,
	we also store an additional macro-transition $\{ s_{t-Nm}, s_{t} \}$ along with its sum of discounted rewards to the shared replay buffer.
	This macro-transition is paired with the current task as its macro-action.
	The Macro-Q Scheduler is then learned with an additional output head attached to $\theta_{\varphi}$ (we also tried $\theta_{\phi}$).
	\item Threshold-Q Scheduler:
	Selects task according to the Q-value output of the extrinsic task head.
	For this scheduler no additional learning is needed.
	It just selects a task based on the current Q-value of the extrinsic head $\theta_e$.
	We tried the following selection strategies:
	\begin{itemize}
		\item Running mean:
		select intrinsic when the current Q-value of the extrinsic head is below its running mean, extrinsic otherwise
		\item Heuristic median:
		observing that the running mean of the Q-values might not be a good statistics for selecting tasks due to the very unevenly distributed Q-values across the map,
		we choose a fixed value that is around the median of the Q-values ($0.007$),
		and choose intrinsic when below, extrinsic otherwise
	\end{itemize}
\end{itemize}



As we report in the paper,
none of the above scheduler choices consistently performs better across all environments than a random scheduler.
We leave this part to future work.

\end{document}